\documentclass[11pt]{article} 

\usepackage{graphicx}
\usepackage{booktabs}
\usepackage{algorithm}
\usepackage{algpseudocode}
\usepackage[T1]{fontenc}
\usepackage{lmodern}
\usepackage[flushleft]{threeparttable}
\usepackage[hidelinks]{hyperref} %
\hypersetup{hidelinks}
\usepackage{multirow}
\usepackage{longtable}
\usepackage[margin=3cm]{geometry} %
\usepackage{amsmath}
\usepackage{authblk}
\usepackage{caption}
\usepackage{cite}

\DeclareMathOperator\erf{erf}

\title{NIDS Neural Networks Using Sliding Time Window Data Processing with Trainable Activations and its Generalization Capability}
\author[1,*]{Anton Raskovalov}
\author[1]{Nikita Gabdullin}
\author[1]{Ilya Androsov}
\affil[1]{Joint Stock ``Research and production company ``Kryptonite'' \authorcr
E-mail: a.raskovalov@kryptonite.ru, n.gabdullin@kryptonite.ru, i.androsov@kryptonite.ru}
\affil[*]{corresponding author}
\date{}
\begin{document}

    \captionsetup[table]{labelformat={default},labelsep=period,name={Table}}

    \maketitle

    \begin{abstract}
        This paper presents neural networks for network intrusion detection systems (NIDS), that operate on flow data preprocessed with a time window. 
        It requires only eleven features which do not rely on deep packet inspection and can be found in most NIDS datasets and easily obtained from conventional 
        flow collectors. The time window aggregates information with respect to hosts facilitating the identification of flow signatures that are missed 
        by other aggregation methods. Several network architectures are studied and the use of Kolmogorov-Arnold Network (KAN)-inspired trainable activation 
        functions that help to achieve higher accuracy with simpler network structure is proposed. The reported training accuracy exceeds 99\% 
        for the proposed method with as little as twenty neural network input features. This work also studies the generalization capability of NIDS, 
        a crucial aspect that has not been adequately addressed in the previous studies. The generalization experiments are conducted using CICIDS2017 dataset 
        and a custom dataset collected as part of this study. It is shown that the performance metrics decline significantly when changing datasets, and the 
        reduction in performance metrics can be attributed to the difference in signatures of the same type flows in different datasets, which in turn can be 
        attributed to the differences between the underlying networks. It is shown that the generalization accuracy of some neural networks can be very unstable 
        and sensitive to random initialization parameters, and neural networks with fewer parameters and well-tuned activations are more stable and achieve higher 
        accuracy.  
    \end{abstract}

    
    \section{Introduction}\label{introduction}

In today's digital landscape, where cyberattacks are increasingly sophisticated and frequent, the need for robust security measures is greater than ever. 
Network Intrusion Detection Systems (NIDS) have emerged as a critical component in safeguarding networks from potential threats. NIDS are designed to monitor 
network traffic for signs of malicious activity, providing a crucial layer of defense against unauthorized access and attacks. By analyzing network packets in 
real-time and using advanced algorithms to detect anomalies and suspicious behavior, NIDS help organizations to quickly identify and respond to threats before 
they can cause significant damage. As cyber threats continue to evolve NIDS become increasingly indispensable.

Advancements in computer science have led to the integration of machine learning into Network Intrusion Detection Systems, significantly enhancing their effectiveness. 
Traditional NIDS rely on predefined signatures and heuristic rules to identify threats, which can limit their ability to detect new or evolving attacks. 
However, Machine Learning (ML) can help NIDS to analyze vast amounts of network traffic and identify patterns that might be indicative of previously unknown threats. 
In addition, ML algorithms can learn and adapt over time. To date there are many papers where ML is used for the purposes of NIDS: Extra Trees~\cite{ref1}, 
fully-connected neural networks~\cite{ref2}, convolution and recurrent neural networks~\cite{ref3,ref4}, graph neural networks (GNN)~\cite{ref5,ref6,ref7,ref8}. 
The popularity of graph-based networks 
for NIDS can be explained by the fact that numerous network hosts and the flows between them can naturally be represented as nodes and edges of a graph, respectively. 

Whereas in our previous work promising results with GNN-based NIDS were achieved~\cite{ref8}, GNNs were found to be very unstable in generalization scenarios 
discussed in Section~\ref{gen-exp}. Graphs can also add unnecessary complexity to the traffic data collection and storage problem. That happens because for datasets 
collected over prolonged periods of time we receive a large number of edges for each node, many of which are temporally distant and should not be considered 
for decision-making. This also creates a problem of selecting the appropriate time frame for dataset collection and the adequacy of transferring learning from 
one graph to another collected over a different time period. Furthermore, the graph topology for NIDS is not very relevant: in real scenarios we know nothing 
about external hosts except that they exchange information with our hosts. We have no information about the data exchange between external hosts, so we can only 
know the edges involving “our” nodes. Thus, the graph degenerates into a ``bristle'' pointing towards external hosts. Message passing along such graphs is not interesting.

For these reasons we have moved away from graph neural networks and adopted a time window-based approach as a faster and more transferable solution that accounts 
for the temporal characteristics of network traffic. To make decisions about flows we collect data from the hosts over a specified time interval. It is worth 
noting that attacks such as DoS, DDoS, and PortScan have very distinct temporal signatures — these always involve a large number flows per unit of time, which 
is not considered in graph-based solutions. From a computational perspective, efficiency improves for a time window-based approach because there is no need 
to build and maintain a graph in memory. We do not need to wait for another flow on the same host; we can proceed with the results immediately. 

Another issue addressed in this paper is the transfer ability of learning across different datasets, or neural network generalization capability. 
Good generalization is desirable meaning that the method should perform well on different datasets without neural network retraining. Prior research 
addressing issues of working with different datasets does not present generalization experiments since all test results were demonstrated on the same dataset 
the networks were trained on~\cite{ref1,ref4,ref5,ref6,ref7}. Moreover, some studies have used different feature sets for different datasets~\cite{ref5,ref6}. 
In this study we propose a 
generalized feature set that we use to access the generalization across datasets. The results are somewhat underwhelming; in some cases it is challenging 
to achieve not only good transfer accuracy but even transfer stability. This issue is discussed in detail in the corresponding section with an explanation of 
the underlying reasons.

The rest of the paper is organized as follows: datasets and their preprocessing is described in Section~\ref{NIDS-datasets}, our method is proposed in 
Section~\ref{methodology}, the obtained results are discussed in Section~\ref{results-discussions}, and Section~\ref{conclusions} concludes the paper.

\section{NIDS datasets}\label{NIDS-datasets}

Whereas various open-access NIDS datasets are available for neural network training~\cite{ref9,ref10,ref11,ref12,ref13,ref14,ref15,ref16},
there is no consistency between data fields is these datasets. 
Moreover, same fields in different datasets might have different formats or be collected under different conditions complicating the use of the datasets. 
Table~\ref{tab:1} illustrates this problem.

\begin{table}
  \caption{Differences between NIDS datasets.}\label{tab:1}
  \centering
  \begin{tabular}{|c|c|c|c|}
    \hline
    \textbf{parameter / dataset} & \textbf{ToN-IoT \cite{ref13}} &
    \textbf{CICIDS2017 \cite{ref14}} & \textbf{own (see Section~\ref{own-dataset})} \\
    \hline
    units of duration & ms & µs & s \\
    \hline
    transferred bytes do /  & both & with headers & without headers \\
    do not include headers & available & &  \\
    \hline
    collector timeout & unspecified & 2 min & 5 min \\
    \hline
    TCP flows are merged & yes & no & yes \\
    \hline
  \end{tabular}
\end{table}

\subsection{CICIDS2017 dataset}\label{CICIDS2017}

In this study we mainly use CICIDS2017 (CDS)~\cite{ref14} dataset for neural network training, which is collected by Canadian Institute for Cybersecurity. 
It consists of raw data pcap files and csv files with aggregated flow data. Unfortunately, we were unable to use CICIDS2017 “as is” due to several issues. 
First, units of some\ fields in CICIDS2017 were different from those obtained from a collector used to produce our dataset (described in the next section), 
and headers were not considered when calculating transferred bytes. Second, the collector timeout was very short making some flows appear deceptively short. 
Most importantly, flows longer than the collector timeout were not merged after the dump in CICIDS2017, which produced significant aggregation artifacts.

To avoid these issues, we used the original pcap files to generate flow data compatible with other datasets using Tranalayzer2~\cite{ref17} with settings PACKETLENGTH=2
(with headers), FLOW\_TIMEOUT=300 (5 min collector timeout), IPV6\_ACTIVATE=0 (ipv4 only).

We used timestamp and src/dst ip addresses to correlate the original CICIDS2017 labels with the new flows. The correctness of such labeling was verified 
by randomly choosing data in pcaps using tshark~\cite{ref18}, checking time and ip addresses and comparing the labels. We have also merged similar CDS classes, such 
as different DoS or PortScan types, and removed classes with few samples resulting in a dataset we labeled as CDSR with six classes: 1475310 benign, 167015 DoS, 
94614 DDoS, 7750 Password, 216866 PortScan, and 661 XSS.

\subsection{Own Dataset}\label{own-dataset}

To test generalization capabilities of neural networks trained on CDSR a small dataset consisting of NetFlowv5 data was collected on our corporate network. 
Benign flows were collected over the duration of two hours during normal weekday working hours to correctly represent realistic diverse normal traffic. 
The network primarily consisted of Windows and Linux machines. During the same time frame PortScan and DoS attacks were conducted in an isolated laboratory 
network inside the main corporate network. 

A laptop with Ubuntu 22.04 was used as attack station to scan ports of other Ubuntu 20.04 Laptops and Ubuntu 20.04 servers. For DoS attack a web server launched 
on one of the servers was used as the target. No Windows machines were used during the attack experiments. Nmap was used for PortScan attacks with and without sudo, 
and flags -sS, -sP, -PA, -PU. Total PortScan duration was one hour. For DoS 20 min of slowhttptest~\cite{ref19} and 40 min of GoldenEye~\cite{ref20} attacks were conducted. 

In total three datasets were collected: 945933 benign + 28244 DoS (referred to as OD);
590238 benign + 18332 PortScan (referred to as OS); and their combination resulting in 1536171 benign + 28244 DoS + 18332 PortScan (referred to as ODS).

\section{Methodology}\label{methodology}

\subsection{Time-Window Processing of Datasets}\label{time-window-proc}

NIDS datasets are commonly distributed as *.csv files of tables with rows representing flows and columns with fields containing flow attributes. 
Assuming that these data are sorted by timestamp, the following algorithm can be used to process datasets using a sliding time window. 

We propose to use the following data structures to aggregate and store flow data. They include three lists for each host that will include flows 
categorized by protocol: TCP, UDP, and Other. For each flow we first check its source and destination IP addresses. When those are encountered for the 
first time, new data structures for source and destination hosts are created. When the data structure for encountered hosts already exists, a time window 
of set length that uses current flow's timestamp as an anchor is used to check which flows from the host's lists are in the time window. All flows that are 
not in the window are discarded. Next, it is checked whether the source and destination ports of the current flow have been used before for the respective 
hosts to set corresponding flags, and the aggregation information for the existing host records is updated. The current flow is added to the appropriate 
host list (TCP, UDP, or Other) based on its protocol. All received information is saved: the flow data, the aggregation information about hosts, and the 
quantitative information about hosts (number of flows and ports of each type), see Algorithm~\ref{alg:1}. After this conversion, the flow and host information for 
the given timestamp is ``frozen''. All necessary data are saved, eliminating the need for further computations or specific processing order. It is possible 
to train on batches of such data randomly sampled from the converted dataset. 

It should be noted that when network activity on a host is high (many flows fall within the time window), calculating aggregation information can be 
time-consuming as it requires recalculating the average over a large number of flows. This problem is often solved by choosing a fixed number of random 
flows for aggregation which adds uncertainty into the analysis. However, in our latest models we decided not to use the information about average values 
of parameters for decision-making and only considered the number of flows and the number of used ports within the time window. These quantities can be 
calculated sufficiently fast even for real-time scenarios. The proposed method is also convenient if there is a need to add additional host metrics such 
as CPU load, disk access, etc. However, studying such metric was outside of the scope of this study.

\begin{algorithm}
  \caption{Flow data processing with a time window}\label{alg:1}
  \begin{algorithmic}[1]
  \State \textbf{initialize} hosts ← \{\}, index ← 0
  \State \textbf{allocate} FLOW, AGGR1, AGGR2, CNT1, CNT2
  \For{rows \textbf{in} dataset}
    \State data1 ← duration, src packets, dst packets, src bytes, dst bytes
    \State data2 ← duration, dst packets, src packets, dst bytes, src bytes

    \If{src\_ip \textbf{not in} hosts}
      \State hosts[src\_ip] ← \textbf{new} host
    \Else
      \State remove flows that are outside the time window from hosts[src\_ip]
    \EndIf

    \If{dst\_ip \textbf{not in} hosts}
      \State hosts[dst\_ip] ← \textbf{new} host
    \Else
      \State remove flows that are outside the time window from hosts[dst\_ip]
    \EndIf

    \State Aggregate data and make counts for hosts[scr\_ip]
    \State Aggregate data and make counts for hosts[dst\_ip]
    \State new\_port1 ← src\_port not in hosts[src\_ip]
    \State new\_port2 ← dst\_port not in hosts[dst\_ip]
    \State FLOW[index] ← flow data, new\_port1, new\_port2
    \State AGGR1[index] ← hosts[src\_ip] aggregated data
    \State AGGR2[index] ← hosts[dst\_ip] aggregated data
    \State CNT1[index] ← counts for each protocol for hosts[src\_ip]
    \State CNT2[index] ← counts for each protocol for hosts[dst\_ip]
    \State add data1 to hosts[src\_ip]
    \State add data2 to hosts[dst\_ip]
    \State index ← index + 1
  \EndFor
  \end{algorithmic}
\end{algorithm}

\subsection{Feature Engineering}\label{feat-eng}

There are two types of dataset features: the raw dataset parameter values obtained from the collector and the derived features added before feeding data 
into the neural network. Typically, host IP addresses are not used directly, but they may be utilized for constructing a graph~\cite{ref5,ref6,ref7,ref8}; 
some parameters 
like protocol are converted into one-hot feature vectors. So, there is no one-to-one relation between these types of features. We will refer to the first 
type of features as ``fields'' and the second one as ``features''. Our core principle is to avoid using fields that are difficult to obtain or for which data 
are missing in well-known datasets. Therefore, all our features are based on universal fields that are available in almost all datasets: duration, packet count, 
byte count, protocol and port numbers, without requiring information obtained by deep packet inspection. This approach also allows us to use the same model with 
the same input dimensions across different datasets, unlike~\cite{ref3}, where a unique set of features was defined for each dataset. Also, note that using a 
large number of fields may lead to overfitting. The set of fields required for our method and their purposes are summarized in Table~\ref{tab:2}.

\begin{table}
  \caption{Necessary dataset fields and their use in time window calculations}\label{tab:2}
  \centering
  \begin{tabular}{|c|c|}
    \hline
    timestamp & Merging flows for hosts \\ \hline
    src ip & Merging flows for hosts \\ \hline
    src port & Counting unique «in use» ports for hosts, evaluating \\
    &  if
    current flow's port have already been in used for the
    host \\ \hline
    dst ip & Merging flows for hosts \\ \hline
    dst port & Counting unique «in use» ports for hosts, evaluating \\ 
    &  if
    current flow's port have already been in used for the
    host \\ \hline
    protocol & Categorizing flows and choosing neural network \\ \hline
    duration & Feature calculation (3 features) \\ \hline
    src packets & Feature calculation (6 features) \\ \hline
    dst packets & Feature calculation (6 features) \\ \hline
    src bytes (with headers) & Feature calculation (6
    features) \\ \hline
    dst bytes (with headers) & Feature calculation (6
    features) \\ \hline
  \end{tabular}
\end{table}

In most cases, the src and dst prefixes do not truly identify which host initiates the exchange since such marking is determined by the first packet that 
arrives within the flow collector's time window. Therefore, field pairs such as (src bytes, dst bytes) are instead ordered by magnitude

\begin{equation}
  \mathrm{feature_{1}, feature_{2}} = \max(\mathrm{src\_bytes, dst\_bytes}), \min(\mathrm{src\_bytes, dst\_bytes}),
  \label{eq:1}
\end{equation}

which makes our neural network invariant to the identification of src and dst hosts. Features used here are listen in Table~\ref{tab:3}.

\begin{table}
  \caption{NIDS Features constructed in this paper.}\label{tab:3}
  \begin{tabular}{|c|l|}
    \hline
      & Features obtained directly from flow fields: \\ \hline
    1 & duration \\ \hline
    2 & abs(src packets -- dst packets)\textsuperscript{a} \\ \hline
    3 & abs(src bytes -- dst bytes)\textsuperscript{b} \\ \hline
    4 & duration / (src bytes + dst bytes) \\ \hline
    5 & duration / (src packets + dst packets) \\ \hline
    6 & max(src packets, dst packets) \\ \hline
    7 & min(src packets, dst packets) \\ \hline
    8 & max(src bytes, dst bytes) \\ \hline
    9 & min(src bytes, dst bytes) \\ \hline
    10 & max(src bytes / (src packets + 10\textsuperscript{-4}), dst bytes /
    (dst packets + 10\textsuperscript{-4}))\textsuperscript{c} \\ \hline
    11 & min(src bytes / (src packets + 10\textsuperscript{-4}), dst bytes /
    (dst packets + 10\textsuperscript{-4})) \\ \hline
      & Features obtained with aggregation information inside current time
    window: \\ \hline
    12 & 0.5 (new port 1 + new port 2)\textsuperscript{d} \\ \hline
    13 & max(src flow count, dst flow count) \\ \hline
    14 & min(src flow count, dst flow count) \\ \hline
    15 & max(src port count, dst port count) \\ \hline
    16 & min(src port count, dst port count) \\ \hline
    17 & abs(src flow count -- dst flow count) \\ \hline
    18 & abs(src port count -- dst port count) \\ \hline
    19 & max(src port count / (src flow count + 10\textsuperscript{-4}), dst
    port count / (dst flow count + 10\textsuperscript{-4})) \\ \hline
    20 & min(src port count / (src flow count + 10\textsuperscript{-4}), dst
    port count / (dst flow count + 10\textsuperscript{-4})) \\ \hline
  \end{tabular}
  \textsuperscript{a} packets refers to the number of packets, \\
  \textsuperscript{b} bytes refers to the number of transferred bytes (with headers), \\
  \textsuperscript{c} adding 10\textsuperscript{-4} to the denominators is done for numerical stability to avoid division by zero, \\
  \textsuperscript{d} new port shows that port of this flow has not been used on the host before in this time window. 
\end{table}

\begin{figure} 
  \centering
  \includegraphics[width=\textwidth]{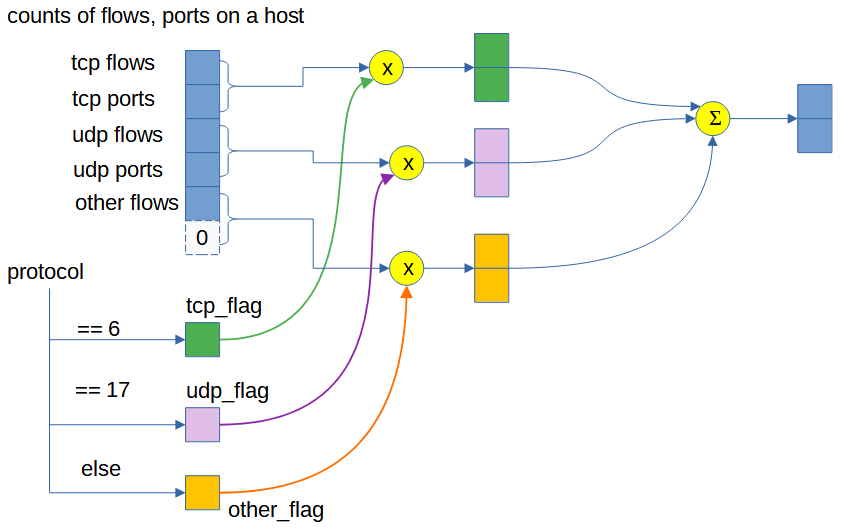}
  \caption{Choosing quantitative host data depending the flow protocol.}\label{fig:1}
\end{figure}

As noted earlier, the data for the time window separately tracks the number of TCP, UDP, and other types of flows as well as the number of ports 
(with TCP and UDP ports counted, and zero for other protocols). All counts (ports and flows) are computed for the protocol to which the flow belongs, as 
illustrated in Figure~\ref{fig:1}.

\subsection{Feature normalization with trainable activation functions}\label{feat-norm-train-activ}

The values of obtained features are generally not bound to a specific range, but for stable neural network performance it is preferable to use 
ranges of -1 to 1 or 0 to 1. The standard practice is to perform feature normalization using the mean and variance of the data~\cite{ref2,ref3,ref5,ref7} or 
feature scaling based on the range of values~\cite{ref1}. Note that normalizing using standard deviation is unfeasible for non-normal distributions and 
can make the model overly dependent on the data distribution in the training dataset. Additionally, these types of normalization still do not strictly 
limit the resulting values that may still fall outside the -1 to 1 interval when transitioning between different datasets. Therefore, instead of normalization, 
we employed learnable activation functions inspired by Kolmogorov-Arnold Networks (KANs)~\cite{ref21}.

One of such functions is a step function, or a ``ladder'' of step functions, which is guaranteed to return a value between zero and one:

\begin{equation}
  y = \frac{1}{n} \sum\limits_{i=1}^{n} \erf(k_{i}(x - x_{i})),
  \label{eq:2}
\end{equation}

where \textit{n} is a number of steps which is a fixed parameter; \textit{k\textsubscript{i}}, \textit{x\textsubscript{i}} are trainable parameters for slopes 
and positions of \textit{i\textsuperscript{th}} step, respectively. Another one is a function containing multiple peaks similar to spline activations in 
KAN~\cite{ref21}:

\begin{equation}
  y = \sum\limits_{i=1}^{n} \exp\left(\frac{-{(x - x_{i})}^2}{w_{i}}\right),
  \label{eq:3}
\end{equation}

where \textit{n} is a number of peaks, fixed parameter; \textit{w\textsubscript{i}}, \textit{x\textsubscript{i}} are trainable parameters for widths and positions 
of peaks, respectively. This function can return a value 
greater than 1 when peaks are close together. In order to avoid steps and peaks getting too close to one another during training, we allow only the ones with the 
least absolute value of their gradient to be modified by setting gradients of the other ones to zero, thus localizing learning, similar to KAN splines~\cite{ref21}. 
The initial parameters of trainable activation functions are not initialized randomly but are chosen considering feature trends in specific datasets for faster training.

\subsection{Neural Network Architectures}\label{nn-arch}

After computing all the features and applying activation functions to them, the results are fed into a neural network with a very simple architecture 
(referred as TWNet). We tested fully connected neural networks with two hidden layers and ReLU activations referred to as ``FF'' in Table~\ref{tab:4}.
Values of \textit{x} and \textit{y} in FF\{x,y\} show the number of neurons in the first and the second layer, respectively. 
We also tested neural networks with no hidden layers with features passed directly to classifier. 

\begin{table}
  \caption{Studied neural network architectures.}\label{tab:4}
  \centering
  \begin{tabular}{|l|l|l|l|l|}
    \hline
    model & number of & features' & activation & layers \\ 
    & features & numbers (see~\ref{feat-eng}) & functions & \\ \hline
    TWNet1 & 7 & 1,3,5,13,15,    & erf (k * x) &  FF\{16,32\} +   \\ 
    & & 2, zero-leng flag & none &  classifier *  \\ \hline
    TWNet2 & 15 (features 6--9  & 1,3,5,13,15 & erf (k * x) &  FF\{16,32\} + \\ 
    & used twice with  & 6--9 (twice)   & erf (x – x\textsubscript{0}) & classifier \\ 
    & different \textit{x\textsubscript{0}} in  & 12 &  none & \\ 
    & activation) & 1 & & \\ \hline
    TWNet3 & 18 & 1 & (\ref{eq:2}), n=3 & classifier /  \\ 
    & & 3,5,13--20 & (\ref{eq:2}), n=1 & FF\{32,32\} +   \\ 
    & & 6--11 & (\ref{eq:2}), n=2 &  classifier \\ 
    & & 12 & none &  \\ \hline
    TWNet4 & 18 & 1 & (\ref{eq:2}), n=3 & classifier /  \\ 
     & & 3,5,13--20 & (\ref{eq:2}), n=1 & FF\{32,16\} +  \\ 
     & & 6--11 & (\ref{eq:3}), n=2 & classifier \\ 
    & & 12 & none & \\ \hline
    TWNet5 & 20 & 1 & (\ref{eq:2}), n=3 & FF\{32,16\} +  \\ 
     & & 2--5,13--20 & (\ref{eq:2}), n=1      & classifier \\ 
     & & 6--11 & (\ref{eq:3}), n=3  & \\ 
    & & 12 & none & \\ \hline
  \end{tabular}
\end{table}

\begin{figure} 
  \centering
  \includegraphics[scale=0.5]{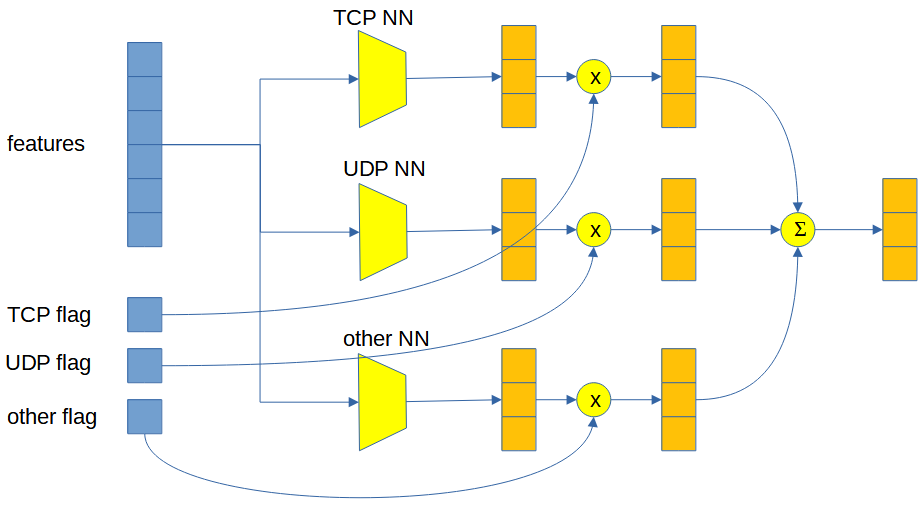} 
  \caption{Applying different neural networks depending on flow's protocol. Since for every flow only one protocol flag is non-zero, only 
  corresponding protocol's neural network output is non-zero, too.}\label{fig:2}
\end{figure}

The fully connected neural networks were designed so that their parameters depended on the flow's protocol (TCP, UDP, or other). 
This was achieved by creating three separate sub networks, one for each protocol, and applying appropriate masks, as shown in Figure~\ref{fig:2}.

\subsection{Training and verification details}\label{train-details}

Training and validation experiments were conducted using two distinct datasets. One dataset (referred to as training dataset) was split to 80/20 subsets with 
the model trained on the 80\% subset. 
After each epoch performance metrics were calculated using the entire second dataset (referred to as testing dataset) with the accuracy value on the second 
dataset referred to as generalization (or testing) accuracy. At the end of training, metrics were also calculated using the entire training dataset.
We used the AdamW optimizer with weight decay set to 10\textsuperscript{-5} and learning rate of 5\texttimes10\textsuperscript{-4}, and cross-entropy as our loss 
function. The neural network trained efficiently, achieving high accuracy (up to 99\%) on the training dataset. Only the multiclass classification task was studied.

\section{Results and Discussion}\label{results-discussions}

\subsection{Neural network generalization}\label{gen-exp}

\begin{figure} 
  \centering
  \includegraphics[scale=0.179]{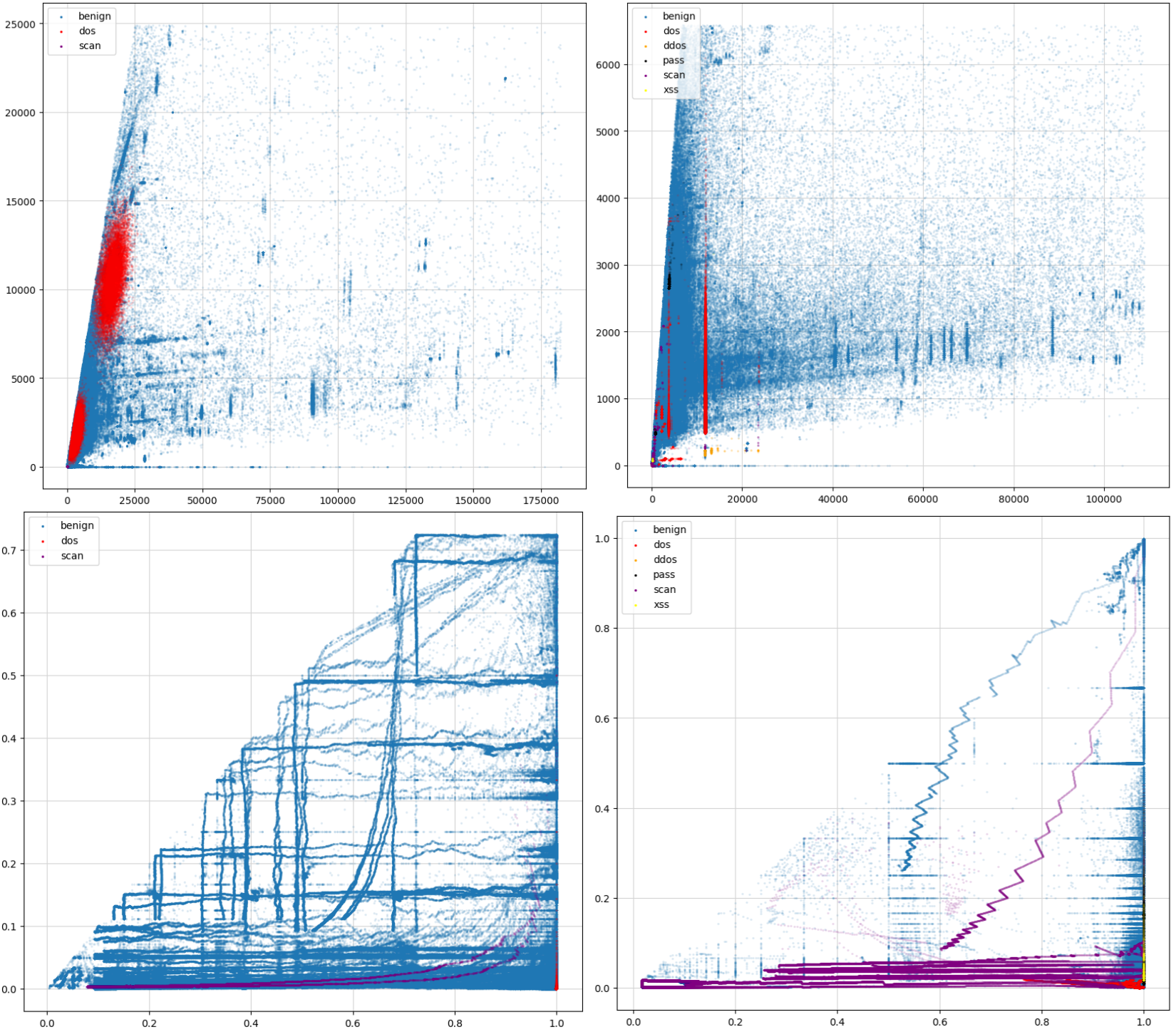}
  \caption{Visualization of value distribution for some data fields in datasets (CDSR and ODS): min vs max numbers of transferred bytes (upper row), 
  and min vs max count of ports per counts of flows between hosts (bottom row).}\label{fig:3}
\end{figure}

The generalization capabilities were studied for neural networks with architectures summarized in Table~\ref{tab:4}. FF GNN in Table~\ref{tab:5} refers 
to GNN used in our previous work~\cite{ref8}.
The effects of random initialization of weights and other parameters were studied by conducting several identical runs with different random seeds 
for both training and testing datasets. It should be stressed again that testing was performed on our dataset, which was different from the training dataset, 
and not on its validation subset. The results are shown in Table~\ref{tab:5} where training accuracy refers to the accuracy calculated for full training dataset
(CDSR, train and validation subsets) after several epochs (see column 3). Testing dataset metrics were calculated on one of subsets of our dataset (OD, OS or ODS) 
without any fine-tuning or retraining of the model. Testing dataset metrics in columns 4--7 include recall (ratio of true positive (TP) for attack type to the 
number of all attacks of that type) for DoS and PortScan (if applicable), correct attack detections rate, and accuracy. 

The correct attack detections rate (CAD) is determined as follows:

\begin{equation}
  CAD = \frac{\sum TP}{\sum (TP + FP)},
  \label{eq:cad}
\end{equation}

for all attacks in dataset.

These metrics for training dataset are shown in later sections. Typically, these metrics were high for well-represented classes (Benign, DoS, DdoS, PortScan) 
and low for classes with fewer samples (Password, XSS).

Table~\ref{tab:5} shows that training dataset accuracy increases and its variation decreases with additional features. It should be stressed that adding features does not 
require additional dataset fields. However, testing dataset metrics were varying drastically between runs even when training accuracy remains the same, with the 
variations sometimes reaching several orders of magnitude. This effect was first observed for the GNN model studied in~\cite{ref8}. 
These variations become less prominent 
for architectures with fewer parameters (without hidden layers), e.g. TWNet4\{0\}. For such models training with different random seeds yields almost the same weights 
with very low cross-run variance and shows almost the same testing accuracy. On the other hand, such architectures are harder to train and they show almost no 
improvements for longer training (higher number of epochs). 

Generalization (testing) accuracy was calculated only for attacks present in both training and testing datasets, i.e. DoS and PortScan. Table~\ref{tab:5} shows 
that PortScan 
generalize much better than DoS. Moreover, in some cases PortScan training and generalization accuracies are almost the same when switching from CDSR dataset to ours,
with recall above 99\%. Unfortunately, this also results in high false positive (FP) rate illustrated by low values in correct attack detections rate (column 6).

These effects can be explained if feature value distributions for training and testing datasets are compared side-by-side. Figure~\ref{fig:3} shows that both attack and 
benign traffic features are very different. 

\begin{table}
  \caption{Generalization experiment results for CDSR as training and ours are testing datasets.}\label{tab:5}
  \centering
  \begin{tabular}{|c|c|c|c|c|c|c|}
    \hline
    \textbf{model} & \textbf{run} & \textbf{accuracy } & \textbf{DoS recall} & \textbf{SCAN recall}
    & \textbf{CAD}
    & \textbf{accuracy} \\
    & & \textbf{on entire} & \textbf{on testing} & \textbf{on testing} & \textbf{on testing} & \textbf{on testing} \\
    & & \textbf{training} & \textbf{dataset, \%} & \textbf{dataset, \%} & \textbf{dataset, \%} & \textbf{dataset, \%} \\
    & & \textbf{dataset, \%} & & & & \\
    \hline
    FF GNN~\cite{ref8} & 1 & 97.0 & - & 33.6 & 66.6 & 96.5 \\
    & 2 & 98.0 & - & 8.3 & 79.0 & 96.5 \\
    & 3 & 97.4 & - & 16.6 & 30.0 & 95.8 \\
    & 4 & 98.7 & - & 0.0 & 0.0 & 73.2 \\ \hline
    TWNet1\{16,32\} & 1 & 93.3 & - & 89.5 & 11.5 & 79.0 \\
    & 2 & 96.6 & - & 93.9 & 5.7 & 53.3 \\
    & 3 & 94.0 & - & 54.1 & 7.3 & 78.1 \\
    & 4 & 96.5 & - & 90.0 & 6.1 & 58.3 \\ \hline
    TWNet2\{16,32\} & 1 & 98.1 & - & 93.3 & 15.1 & 83.9 \\
    & 2 & 95.8 & - & 99.9 & 17.0 & 85.3 \\
    & 3 & 95.7 & - & 94.7 & 26.2 & 91.8 \\
    & 4 & 95.8 & - & 9.0 & 2.9 & 88.3 \\ \hline
    TWNet2\{16,32\} & 1 & 95.7 & 0.8 & - & 0.2 & 87.5 \\
    & 2 & 98.0 & 20.0 & - & 3.0 & 79.1 \\
    & 3 & 95.8 & 0.4 & - & 0.1 & 87.0 \\
    & 4 & 95.8 & 0.9 & - & 0.3 & 87.7 \\ \hline
    TWNet3\{0\} & 1 & 98.6 & 69.9 & 99.1 & 13.6 & 84.2 \\
    & 2 & 98.6 & 20.4 & 98.1 & 11.9 & 87.4 \\
    & 3 & 98.5 & 9.5 & 95.1 & 10.9 & 88.0 \\
    & 4 & 98.7 & 32.2 & 98.1 & 12.4 & 86.7 \\ \hline
    TWNet3\{32,32\} & 1 & 98.9 & 4.6 & 99.7 & 9.9 & 87.1 \\
    & 2 & 98.9 & 32.8 & 99.7 & 26.5 & 93.9 \\
    & 3 & 99.0 & 15.7 & 99.7 & 25.8 & 94.3 \\
    & 4 & 98.8 & 0.0 & 40.1 & 3.1 & 62.5 \\ \hline
    TWNet4\{0\} & 1 & 98.7 & 3.0 & 99.4 & 7.0 & 82.3 \\
    & 2 & 98.6 & 2.8 & 98.9 & 9.3 & 86.6 \\
    & 3 & 98.6 & 2.9 & 99.1 & 9.7 & 87.1 \\
    & 4 & 98.6 & 2.9 & 99.4 & 6.2 & 79.9 \\ \hline
    TWNet4\{32,16\} & 1 & 98.9 & 3.8 & 99.7 & 7.9 & 84.0 \\
    & 2 & 98.9 & 9.8 & 99.7 & 9.1 & 85.1 \\
    & 3 & 98.9 & 12.3 & 99.7 & 4.0 & 65.5 \\
    & 4 & 99.0 & 2.8 & 99.7 & 5.6 & 78.0 \\ \hline
    TWNet5\{32,16\} & 1 & 99.0 & 3.1 & 99.6 & 7.3 & 83.0 \\
    & 2 & 99.0 & 2.8 & 99.6 & 4.0 & 69.4 \\
    & 3 & 98.9 & 15.1 & 99.8 & 5.1 & 72.0 \\
    & 4 & 99.0 & 43.0 & 99.6 & 6.5 & 71.5 \\
    \hline
  \end{tabular}
\end{table}

\subsection{Test of catastrophic forgetfulness and the ability to retrain}\label{test-forget}

One of the reasons that the features in Figure~\ref{fig:3} are very different are the specifics of particular networks, such as network type, collector timeout, 
server types, etc.
In this Section we study the possibility to retrain a neural network model trained on one dataset using another one. We also study the effects of «catastrophic 
forgetfulness», or how the initial training dataset`s accuracy decreases during retraining. It is expected that the use of local learning~\cite{ref21} 
in our model should reduce such effects.
The experiments were organized as follows: TWNet5\{32,16\} was randomly initialized and trained for 8 epochs on the first dataset, with both training and generalization 
accuracies recorded after training is complete. Then the model was trained for 4 epochs on the second dataset witch accuracy recorded for both datasets along with 
additional metrics for the first dataset. Table~\ref{tab:6} summarizes the results for experiments, with first dataset being CDSR and the second one being our dataset. 
Datasets are switched for the experiments summarized in Table~\ref{tab:7}.

\begin{table}
  \caption{Results of retraining from CDSR to ODS datasets for TWNet5\{32,16\}.}\label{tab:6}
  \centering
  \begin{tabular}{|c|c|c|c|c|c|c|c|}
    \hline
    & \multicolumn{2}{l|}{train CDSR (8 epochs),} & \multicolumn{5}{l|}{train ODS (4 epochs), verify CDSR} \\ 
    & \multicolumn{2}{l|}{verify ODS} &  \multicolumn{5}{l|}{} \\
    \hline
    \textbf{run} & \textbf{accuracy} & \textbf{accuracy} & \textbf{accuracy} & \textbf{accuracy} & \textbf{DoS, \%}
    & \textbf{SCAN,\%} & \textbf{CAD, \%} \\
    & \textbf{CDSR, \%} & \textbf{ODS, \%} & \textbf{ODS, \%} & \textbf{CDSR, \%} & \textbf{recall,\%} & \textbf{recall,\%} & \\ \hline
    1 & 99.0 & 83.0 & 99.5 & 81.3 & 0.0 & 58.4 & 94.7 \\
    2 & 99.0 & 77.2 & 99.5 & 80.7 & 0.0 & 52.8 & 94.5 \\
    3 & 99.0 & 69.4 & 99.5 & 81.4 & 0.2 & 59.0 & 94.9 \\
    4 & 98.9 & 72.0 & 99.5 & 81.3 & 0.0 & 58.2 & 94.9 \\
    5 & 99.0 & 81.7 & 99.5 & 81.2 & 0.2 & 57.8 & 94.5 \\
    6 & 99.0 & 71.5 & 99.5 & 81.5 & 0.0 & 60.1 & 94.9 \\
    7 & 98.9 & 82.3 & 99.5 & 81.5 & 0.0 & 59.9 & 95.1 \\
    8 & 99.0 & 73.2 & 99.5 & 81.2 & 0.0 & 58.0 & 94.9 \\
    \hline
  \end{tabular}
\end{table}

\begin{table}
  \caption{Results of retraining from ODS dataset to CDSR datasets for TWNet5\{32,16\}}\label{tab:7}
  \centering
  \begin{tabular}{|c|c|c|c|c|c|c|c|}
    \hline
    & \multicolumn{2}{l|}{train ODS (8 epochs),} & \multicolumn{5}{l|}{train CDSR (4 epochs), verify ODS} \\ 
    & \multicolumn{2}{l|}{verify CDSR} &  \multicolumn{5}{l|}{} \\
    \hline
    \textbf{run} & \textbf{accuracy} & \textbf{accuracy} & \textbf{accuracy} & \textbf{accuracy} & \textbf{DoS, \%}
    & \textbf{SCAN,\%} & \textbf{CAD, \%} \\
    & \textbf{ODS, \%} & \textbf{CDSR, \%} & \textbf{CDSR, \%} & \textbf{ODS, \%} & \textbf{recall,\%} & \textbf{recall,\%} & \\ \hline
    1 & 99.5 & 80.8 & 98.9 & 93.4 & 80.1 & 99.6 & 29.2 \\
    2 & 99.5 & 80.7 & 98.8 & 83.5 & 88.7 & 99.8 & 14.3 \\
    3 & 99.5 & 80.7 & 98.9 & 88.7 & 7.0 & 99.8 & 11.7 \\
    4 & 99.5 & 80.6 & 98.8 & 90.5 & 92.3 & 99.6 & 22.9 \\
    5 & 99.5 & 80.7 & 98.9 & 92.3 & 76.0 & 99.7 & 25.8 \\
    6 & 99.5 & 81.2 & 98.8 & 92.9 & 57.8 & 99.7 & 25.7 \\
    7 & 99.5 & 81.2 & 98.9 & 91.1 & 79.5 & 99.8 & 23.1 \\
    8 & 99.5 & 81.2 & 98.9 & 91.2 & 84.0 & 99.7 & 23.7 \\
    \hline
  \end{tabular}
\end{table}

As it can be expected, the first training dataset accuracy was high in all experiments. However, the model was easily retrained with new data, ``forgetting'' the 
initial training, as illustrated by the reduced accuracy on first dataset with retraining epochs. For instance, CDSR accuracy has reduced from 99\% to 81\% 
with only 50--60\% PortScans correctly detected with low attack FP rate. On the contrary, 
PortScan detection accuracy is high when datasets are switched, but FP rate is high, too. It can be observed that DoS accuracy varies significantly among runs. 
The comparison of columns 1 and 3 in Tables~\ref{tab:6} and~\ref{tab:7} shows that retraining dataset accuracy does not depend on training dataset accuracy. 
It is also shown that the 
accuracy on a dataset is higher if the model was previously trained on that dataset, even if it was consequently retrained on another dataset. The tables also show 
that PortScan generalize better than DoS due to higher similarity of PortScan features in both datasets.

\subsection{Confusion matrix, f1-score}\label{confusion}

Table~\ref{tab:8} shows the confusion matrix on full CDSR dataset for TWNet5\{32,16\} model trained on the training (80\%) subset of CDSR. 
Such matrices were calculated for multiple runs showing low variance and the same characteristic patterns discussed below.

\begin{table}
  \caption{Confusion matrix and metrics for TWNet5\{32,16\} on full CDSR dataset.}\label{tab:8}
  \centering
  \begin{tabular}{|l|c|c|c|c|c|c|c|}
    \hline
    & \textbf{Benign} & \textbf{DoS} & \textbf{DDoS} & \textbf{Password} &
    \textbf{PortScan} & \textbf{XSS} & \textbf{Amounts} \\ 
    \hline
    \textbf{Benign} & 1457391 & 7583 & 1219 & 739 & 8367 & 11 &   1475310 \\
    \hline
    \textbf{DoS} & 188 & 166558 & 79 & 159 & 31 & 0 & 167015 \\
    \hline
    \textbf{DDoS} & 32 & 129 & 94440 & 0 & 0 & 13 & 94614 \\ 
    \hline
    \textbf{Password} & 120 & 0 & 0 & 7600 & 0 & 30 & 7750 \\
    \hline
    \textbf{PortScan} & 543 & 7 & 0 & 1 & 216315 & 0 & 216866 \\
    \hline
    \textbf{XSS} & 23 & 0 & 0 & 631 & 0 & 7 & 661 \\
    \hline
    \textbf{Total Found} & 1458297 & 174277 & 95738 & 9130 & 224713 & 61     & \\
    \hline
    \textbf{True Positive} & 1457391 & 166558 & 94440 & 7600 & 216315 & 7     & \\
    \hline
    \textbf{Recall} & 0.99 & 1.00 & 1.00 & 0.98 & 1.00 & 0.01     & \\
    \hline
    \textbf{Precision} & 1.00 & 0.96 & 0.99 & 0.83 & 0.96 & 0.11     & \\
    \hline
    \textbf{F1-Score} & 0.99 & 0.98 & 0.99 & 0.9 & 0.98 & 0.02     & \\
    \hline
  \end{tabular}
\end{table}

Table~\ref{tab:8} shows that the ration of missed attacks to all attacks is very low, varying between 0.1 and 0.3\% depending on the run, with most missed attacks being PortScans. 
The ratio of FP attacks to all benign is higher, reaching 1.2--1.3\%, with 90\% FP being mislabeled as PortScans and DoS. DDoS is mostly confused with DoS and never 
with Password or PortScan. Similarly, PortScan and Password are never confused with DDoS. PortScan is mostly confused with benign, and XSS is mostly confused with 
Password (95--97\%).

Table~\ref{tab:9} shows f1-scores for each attack type calculated using the confusion matrix for comparison of our results with the results in prior publications. 
The following
abbrebiations were used for datasets: UNSW-NB15 (NB15), BoT-IoT (BoT), ToN-IoT (ToN), UQ-NIDS (UQ), KDD CUP99 (CUP99),  CIC-DarkNet (DN), CSE-CIC-IDS2018 (CSE). 
Prefix ``NF'' means Net FLow feature version of the dataset, ``v2'' means second version of the dataset, ``R'' means rectified version of the dataset. 
In ``features'' column the first number means the number of features used
as input for the ``method'', and the number in parentheses is the number of dataset fields used to calculate the features, as discussed in Section~\ref{feat-eng}. 

The comparison shows that apart from poorly represented attacks, such as XSS, our results are comparable or better than the previously reported ones. In some cases, 
the authors obtained metrics only with a large number of features~\cite{ref1,ref6,ref9}, which may indicate overfitting. Moreover in~\cite{ref5} 
reducing the number of features significantly degrades metrics.

\begin{table}
  \caption{Comparison of f1-score for each attack type for various method and various datasets.}\label{tab:9}
  \begin{tabular}{|l|l|c|c|c|c|c|c|c|}
    \hline
    dataset & method & features & Benign & DoS & DDoS & Pass & Scan & XSS \\ \hline
    NFNB15v2~\cite{ref1} & Extra Trees~\cite{ref1} & 39 (43) & 1.00 & 0.36 & - & - & - & - \\ \hline
    NFBoTv2~\cite{ref1} & Extra Trees & 38 (43) & 1.00 & 1.00 & 1.00 & - & - & - \\ \hline
    NFToNv2~\cite{ref1} & Extra Trees & 38 (43) & 0.99 & 0.91 & 0.99 & 0.97 & 1.00 & 0.96 \\ \hline
    NFUQv2 \cite{ref1} & Extra Trees & 38 (43) & 0.96 & 1.00 & 1.00 & 0.96 & 0.98 & 0.97 \\ \hline
    CUP99~\cite{ref9} & CNN+LSTM~\cite{ref4} & -\textsuperscript{a} & 0.94\textsuperscript{b} & 0.97\textsuperscript{b} & - & - & - &  - \\ \hline
    NSL\_KDD~\cite{ref10} & CNN+LSTM & -\textsuperscript{a} &   0.99\textsuperscript{b} & 0.98\textsuperscript{b} & - & - & - &   - \\ \hline
    NB15~\cite{ref11} & CNN+LSTM & -\textsuperscript{a} &   0.92\textsuperscript{b} & 0.53\textsuperscript{b} & - & - & - &   - \\ \hline
    BoT~\cite{ref12} & e-GraphSage~\cite{ref5} & 47 & 0.99 & 1.00 & 1.00 & - & - & - \\ \hline
    NFBoT~\cite{ref22} & e-GraphSage & 12 & 0.42 & 0.47 & 0.39 & - & - & - \\ \hline
    ToN~\cite{ref13} & e-GraphSage & 44 & 0.91 & 0.73 & 0.98 & 0.91 & 0.85 & 0.95 \\ \hline
    NFToN~\cite{ref22} & e-GraphSage & 12 & 0.92 & 0 & 0.68 & 0.25 & 0.13 & 0 \\ \hline
    NB15~\cite{ref11} & E-ResGAT~\cite{ref6} & 39 (43) & 1.00 & 0.05 & - & - &  - & - \\ \hline
    DN~\cite{ref15} & E-ResGAT & 73 (77) & 0.95 & - & - & - & -  & - \\ \hline
    CSE~\cite{ref16} & E-ResGAT & 73 (77) & 0.98 & 0.96 & 0.99 & - & - & - \\ \hline
    ToN~\cite{ref13} & E-ResGAT & 39 (39) & 1.00 & 0.99 & 0.98 & 1.00 & 0.99 & 1.00 \\ \hline
    BoT~\cite{ref22} & GNN~\cite{ref7} & 39 & 0.58 & 0.31 & 0.46 & - & - & - \\ \hline
    NFBoTv2~\cite{ref1}1 & GNN & 38 (43) & 0.72 & 0.95 & 0.97 & - &  - & - \\ \hline
    NFCSE~\cite{ref22} & GNN & 39 & 0.98 & 0--0.87\textsuperscript{c} & 0-1.00\textsuperscript{c} & \textsuperscript{d} & - & \textsuperscript{d} \\ \hline

    NFCSEv2~\cite{ref1} & GNN & 39 (43) & 0.99 & 0--0.99\textsuperscript{c} & 0.79- & \textsuperscript{d} & - & \textsuperscript{d} \\ 
    & & & & & 0.94\textsuperscript{c} & & & \\ \hline

    NB15~\cite{ref11} & CNN-BiLSTM~\cite{ref3} & 10--100 (41) & 0.93 & 0.19 & - & - & -  & - \\ \hline
    CICIDS2017~\cite{ref14} & CNN-BiLSTM & 10--50 (77) & 1.0 & 1.00\textsuperscript{e} & 1.00\textsuperscript{e} & 0.99\textsuperscript{f} & 1.00  & \textsuperscript{d} \\ \hline   
    CUP99~\cite{ref9} & FF~\cite{ref2} & 125 (41) & 1.00 & 1.00 & - & - & - & - \\ \hline
    ToN-R~\cite{ref8} & GNN~\cite{ref8} & 10 (11) & 0.98 & 0.99 & 0.99 & 0.86 & 0.99 & 0.87 \\ \hline
    
    CDSR & TWNet5\{32,16\} & 20 (11) & 0.99 & 0.98 & 0.99 & 0.88- & 0.98 & 0- \\ 
    (this work) & (this work) & & & & & 0.9 & & 0.03 \\ \hline

    ODS & TWNet5\{32,16\} & 20 (11) & 1.0 & 0.87- & - & - & 1.0 & - \\
    (this work) & (this work) & & & 0.88 & & & & \\ \hline
  \end{tabular}
  \textsuperscript{a} not mentioned, \\
  \textsuperscript{b} reffered as accuracy in multi-classfication, \\
  \textsuperscript{c} different types of this attack accounted separately, \\
  \textsuperscript{d} mixed with other attack types, \\
  \textsuperscript{e} DoS and DDoS combined in one class, \\
  \textsuperscript{f} mentioned as BruteForce, without web variants of passwords attacks.
\end{table}

\newpage

\section{Conclusions}\label{conclusions}

In this paper we discuss the limitation of GNNs for NIDS and propose a sliding time window methodology for data processing and feature extraction. It showed good 
results when combined with simplistic neural network architectures illustrated by over 90\% f1-score for most attacks in CICIDS2017. We propose to use fewer neural 
network input features calculated using a low number of simplest dataset fields available in nearly every NIDS dataset. 

The generalization capabilities of the methods were studied with an emphasis on data compatibility among various datasets. To conduct the generalization experiments 
the CICIDS2017 pcaps were used to create a more compatible version of the dataset, and another dataset with Benign, PortScan, and DoS traffic flows was collected in 
our network. The generalization experiments showed that high generalization accuracy is hard to achieve mainly because of the differences in data patterns in different 
networks due to high variability of settings, including collector type and timeout interval, which makes learning the characteristic features of benign traffic 
extremely challenging. It is also shown that generalization accuracy increases even after a slight fine-tuning on the new dataset, with a detriment to the initial 
training dataset`s accuracy. The highest generalization accuracy was shown for PortScans since these attacks had the most similar signatures in both datasets.

We proposed the use of trainable activation functions that were very effective in separating various features in interpretable manner. Using such functions while 
carefully choosing input features we were able to achieve high training and generalization accuracies with neural networks with few or even no hidden layers. That is, 
we showed that high performance and generalization stability can be achieved by neural networks consisting only of trainable activations and a classifier head. We also 
showed that such neural networks are the least susceptible to uncertainties due to random weight initialization which makes them more reliable in real-world 
applications.

\section*{Acknowledgement}\label{acknowledgement}

Authors would like to thank Vasily Dolmatov for discussions and project supervision.

\section*{Data Statement}\label{Data-Statement}

CDSR dataset is available on request for academic research purposes.


\bibliographystyle{IEEEtran}
\bibliography{IEEEabrv,ms}

\end{document}